\documentclass[runningheads]{llncs}
\usepackage[T1]{fontenc}
\usepackage{graphicx}
\usepackage{tikz}
\usepackage{comment}
\usepackage{amsmath,amssymb} 
\usepackage{color}
\usepackage{cite}
\usepackage{subcaption}
\usepackage{enumitem}
\usepackage[margin=1.4in]{geometry}

\usepackage[accsupp]{axessibility}  

\usepackage{booktabs}
\usepackage{flushend}
\usepackage[pagebackref,breaklinks,colorlinks]{hyperref}

\begin{document}
\title{Unsupervised Structure-Consistent Image-to-Image Translation}
%
%
\author{Shima Shahfar \and
 Charalambos Poullis}
\authorrunning{S. Shahfar and C. Poullis}
%
\institute{
Department of Computer Science and Software Engineering,\\
Concordia University, \\ 
Montreal, Quebec, Canada 
}
\maketitle              
\begin{abstract}
The Swapping Autoencoder achieved state-of-the-art performance in deep image manipulation and image-to-image translation. We improve this work by introducing a simple yet effective auxiliary module based on gradient reversal layers. The auxiliary module's loss forces the generator to learn to reconstruct an image with an all-zero texture code, encouraging better disentanglement between the structure and texture information. The proposed attribute-based transfer method enables refined control in style transfer while preserving structural information \textit{without} using a semantic mask. To manipulate an image, we encode both the geometry of the objects and the general style of the input images into two latent codes with an additional constraint that enforces structure consistency. Moreover, due to the auxiliary loss, training time is significantly reduced. The superiority of the proposed model is demonstrated in complex domains such as satellite images where state-of-the-art are known to fail. Lastly, we show that our model improves the quality metrics for a wide range of datasets while achieving comparable results with multi-modal image generation techniques.

\vspace{-10pt}
\keywords{structure-consistent image-to-image translation \and style transfer \and training class imbalance}
\vspace{-10pt}

\end{abstract}
\vspace{-10pt}
\section{Introduction}
\label{sec:intro}
\vspace{-5pt}

\begin{figure}[!ht]
\begin{center}
\includegraphics[width=\textwidth]{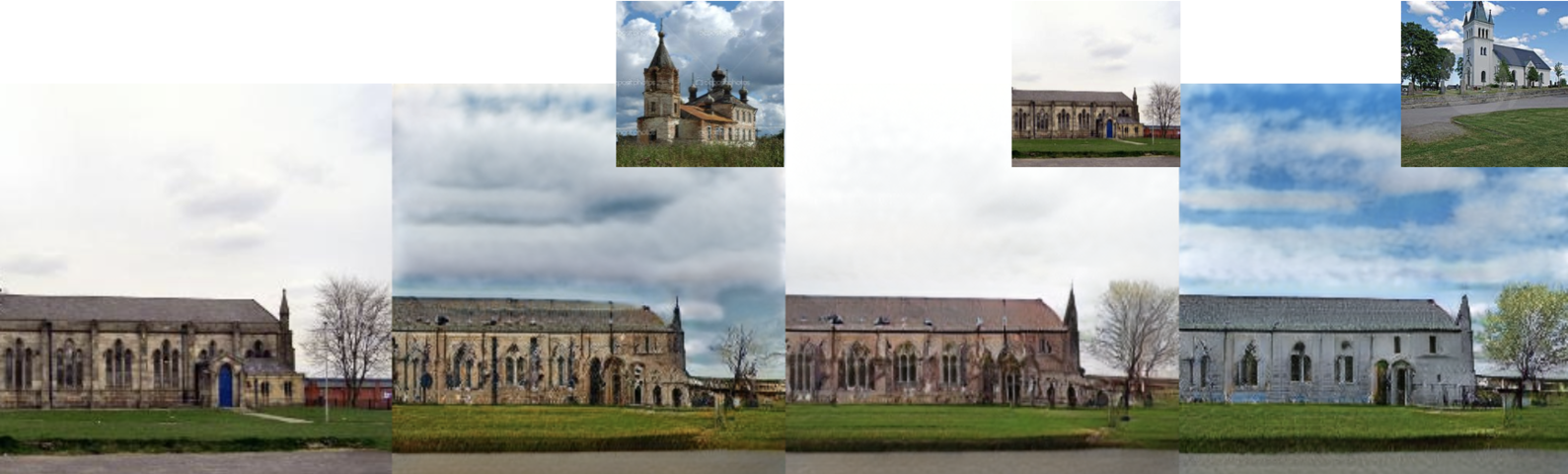}
\end{center}
\caption{
Our method learns structure-consistent image-to-image translation \textit{without} requiring a semantic mask. We learn to disentangle structure and texture for applications such as style transfer and image editing tasks. The first(left) image shows the first input image, and the other images show the generated images in which the structure is retained from the first input image and the texture from the second, third, and fourth input images, respectively, shown in the inset images. Note that the tree’s structure is preserved, and its texture -in this case, the foliage’s colour and density- changes according to the texture of the second input image in the inset. Our model was not trained on any season transfer dataset.
}
\label{fig:main}
\end{figure}

Image-to-image translation and image manipulation techniques attracted much attention \cite{wang2018pix2pixHD, Kaneko2017GenerativeAC, DBLP:journals/corr/LiuBK17, DBLP:journals/corr/YiZTG17, DBLP:journals/corr/IsolaZZE16, CycleGAN2017, DBLP:journals/corr/abs-1711-07410, DBLP:journals/corr/abs-1910-10223, DBLP:journals/corr/abs-1911-11758, karras2020analyzing} recently as they can have a significant effect on many different tasks. Of particular interest is creating realistic synthetic training datasets to improve models' performance and generalization. One example that demonstrates the use of a synthetic dataset in the training of networks is presented in \cite{zhang21} where the authors introduce a semi-supervised approach to generate datasets for semantic segmentation. 

There are a plethora of works \cite{DBLP:journals/corr/abs-1812-04948, karras2020analyzing, Kotovenko_2019_ICCV} which report that for images containing single objects such as faces, or for images having the same semantic layout such as building facades, deep image manipulation techniques can produce realistic synthetic images. However, generating natural scenes or more visually complex images remains a challenge due to differences in the semantic layouts of the input images. 

The challenge of deep image manipulation state-of-the-art with complex scenes is recognizing and learning essential features and characteristics from the input image. Structural information is typically shared or has common characteristics across different images in a dataset. On the other hand, the texture appears entangled with intrinsic image features. The standard approach to preserving the structural information is to condition the generation process on the input semantic mask using conditional image synthesis frameworks. However, that approach is not practical for image manipulation since the assumption of having access to semantic masks does not hold in most cases. Researchers explored different methods such as \cite{DBLP:journals/corr/abs-1811-11155, DBLP:journals/corr/abs-1911-11758}, but in this work, we assume that image representations can be disentangled into the content/structure and texture/style. 

To address this problem, we propose an auxiliary module that enforces the separation of structure from texture. This branch promotes the disentanglement of structure and texture by suppressing texture-related information in the structure code by applying a gradient reversal layer. Additionally, it encourages the emergence of deep features that are highly important for image editing tasks.
Better structure preservation can also impact many applications ranging from creating a 3D synthetic simulation world, image editing, semantic image synthesis, and style transfer. More importantly, the proposed technique can remove biases from training datasets caused by class imbalances. Many benchmark datasets introduce bias \cite{liu2015faceattributes, DBLP:journals/corr/CordtsORREBFRS16} that can limit the generalization capability of any network trained on them and significantly limit the impact of networks trained on these datasets in real-world scenarios. 

\begin{figure}[!ht]
\centering
    \begin{subfigure}{0.47\textwidth}
        \includegraphics[width=\textwidth]{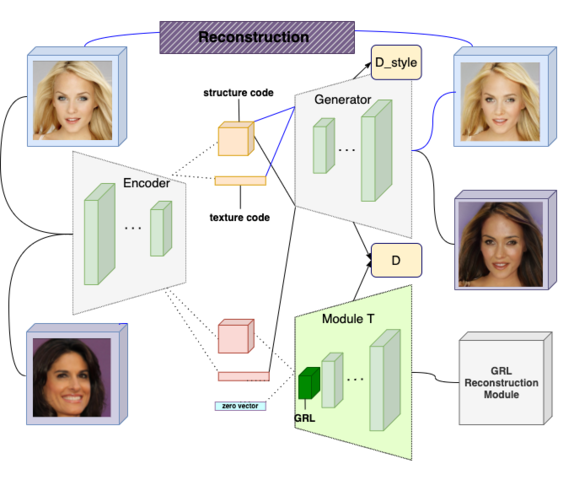}
        \caption{}
        \label{fig:overview}
    \end{subfigure}
    \hfill
    \begin{subfigure}{0.47\textwidth}
        \includegraphics[width=\textwidth]{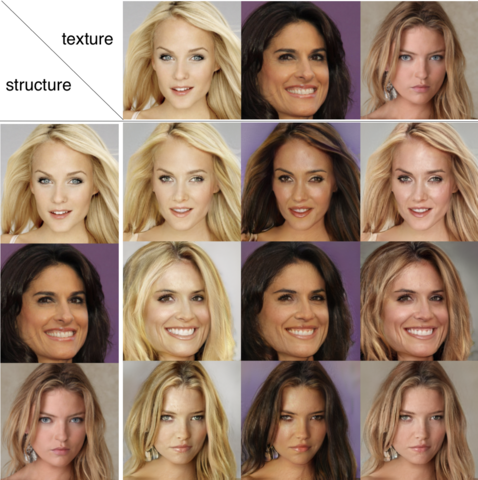}        
        \caption{ }    
        \label{fig:celeba}
    \end{subfigure}
    \caption{(a) \textbf{Overview.} The geometry of the objects and the general style of the input images are encoded into two latent codes with an additional constraint that enforces structure consistency. We introduce a new module that encourages better disentanglement between the structure and the style, based on gradient reversal layers. This results in an attribute-based transfer that allows for a finer style transfer control while preserving structural information without requiring a semantic mask. (b) \textbf{Performance on CelebAMask-HQ:} Our model generates structure-consistent samples while transferring style from one image to another. Unlike most models that fail to preserve small structural details, our approach is able to preserve fine details such as earrings (see last row).}
\end{figure}

This paper pursues three main objectives: 1) consistent and accurate structure preservation, 2) diverse, and 3) realistic image synthesis. Our goal is to learn multi-modal structure-consistent image-to-image translation in a fully unsupervised approach without requiring semantic segmentation masks. Our technical contributions can be summarized as follows:
\vspace{-5pt}
\begin{itemize}
\setlength\itemsep{-0.2em}
\item A new approach for a structure-consistent image-to-image translation that does not rely on prior knowledge on the scene geometry.
\item An auxiliary module that enforces the disentanglement between the structure and texture information with an explicit loss term for penalizing the synthesis of realistic images when no texture information is provided. 
\item An extension of the Swapping Autoencoder model with our auxiliary module. We quantitatively and qualitatively demonstrate that our method generates synthetic images structurally consistent with the source input image. 
\end{itemize}
We present experiments on several datasets, simple datasets with minimal variations in the semantic information of the training examples
such as CelebAMaskHQ \cite{CelebAMask-HQ} Figure \ref{fig:celeba},  and complex datasets where the semantic information varies drastically such as the LSUN Church \cite{yu15lsun} Figure \ref{fig:main}, and Cityscapes \cite{DBLP:journals/corr/CordtsORREBFRS16} Figure \ref{fig:cityscapes}. Our results demonstrate that the proposed method improves the performance at a fraction of the training time required by state-of-the-art.


\vspace{-40pt}
\section{Background and Related work}
 \label{sec:related_work}
\vspace{-5pt}
This section provides an overview of the most relevant state-of-the-art, grouped according to their methodology.

\textbf{Generative models.} Generative Adversarial Networks (GANs) \cite{goodfellow2014generative} introduced an adversarial process to train a generative model. The problem is formulated as a zero-sum game between a generator and discriminator where the optimal solution is to find a Nash equilibrium. Ian J. Goodfellow refers to this framework as a minimax two-player game in which generator G tries to minimize the probability of the discriminator D to recognize the fake samples, and D tries to maximize the probability of assigning the correct label. The objective function is given by,
\begin{equation} \label{eq_gan_basic}
\min \limits_{G} \max \limits_{D} V(D,G) = E_{x \sim p_{data}(x)}[\log D(x)] +E_{z \sim p_z(z)}[\log (1-D(G(z)))]
\end{equation}
GANs have proven to be very successful \cite{ DBLP:journals/corr/ChenDHSSA16, DBLP:journals/corr/ZhuKSE16, DBLP:journals/corr/abs-1812-04948, karras2020analyzing} compared to other common approaches such as \cite{HintonSalakhutdinov2006b, DBLP:journals/corr/OordKVEGK16, DBLP:journals/corr/SalimansKCK17, DBLP:journals/corr/OordKK16, 10.1145/1390156.1390294}. Both GANs and Variational Autoencoders (VAEs) \cite{kingma2014autoencoding} contain an encoder and a decoder; however, they differ in the sense that the GAN is a framework for estimating data distribution. On the other hand, VAEs learn the stochasticity within the data using the encoder's latent code to match the Gaussian distribution by reparameterizing the latent distribution and maximizing the log-likelihood function. Some methods \cite{DBLP:journals/corr/BaoCWLH17, DBLP:journals/corr/abs-1711-11586} combine GAN and VAE or GAN and Autoencoders in their models to achieve multi-modal image generation and prevent mode collapse. 

\textbf{Conditional generative models} such as conditional VAEs \cite{sohn2015learning}, conditional GANs \cite{DBLP:journals/corr/MirzaO14}, conditional autoregressive methods \cite{DBLP:journals/corr/OordKVEGK16, DBLP:journals/corr/GuadarramaDBNS017}, to name a few, have shown promising results \cite{CycleGAN2017} but we focus on conditional GANs for the rest of this section. Generative adversarial networks can be extended to conditional generative models \cite{DBLP:journals/corr/MirzaO14} by feeding additional information $c$ into the discriminator and generator. This $c$ can be any information such as edge mask for semantic segmentation task or class labels for classification. By doing so, the generator can use prior noise $p_z(z)$ and additional information $c$ to create a hidden representation and the discriminator will use the information provided as an input for a better discrimination. The quality of the results generated using conditional GANs inspired many applications employing this method, including, but not limited to, image-to-image translation \cite{wang2018pix2pixHD, Kaneko2017GenerativeAC, DBLP:journals/corr/LiuBK17, DBLP:journals/corr/YiZTG17}, image editing \cite{DBLP:journals/corr/abs-1711-10678, Choi_2018_CVPR}, image inpainting \cite{DBLP:journals/corr/YehCLHD16, DBLP:journals/corr/abs-1711-08590, DBLP:journals/corr/abs-1905-01723}, text-to-image \cite{DBLP:journals/corr/ZhangXLZHWM16, DBLP:journals/corr/abs-1711-10485}, photo colorization \cite{DBLP:journals/corr/ZhangZIGLYE17, DBLP:journals/corr/abs-1906-09909, DBLP:journals/corr/abs-2005-05207, DBLP:journals/corr/SangkloyLFYH16}, conditional domain adaptation \cite{Choi_2018_CVPR, DBLP:journals/corr/BousmalisSDEK16, DBLP:journals/corr/abs-1912-01865, DBLP:journals/corr/abs-1909-07877}, super resolution \cite{DBLP:journals/corr/JohnsonAL16, DBLP:journals/corr/LedigTHCATTWS16}, style transfer \cite{Gatys2016ImageST, DBLP:journals/corr/JohnsonAL16, DBLP:journals/corr/abs-1812-04948, DBLP:journals/corr/HuangB17, karras2020analyzing, xian2018texturegan}. Our work extends the image-to-image translation framework with a focus on image manipulation and style transfer.  

\textbf{Image-to-image translation} is a framework to transfer an input image into a synthesized output image while preserving some information from the input. There are many methods designed for different applications. The main difference is in the information they preserve from the input image, which depends on the application. 
Image-to-image translation showed promise \cite{DBLP:journals/corr/IsolaZZE16, CycleGAN2017, DBLP:journals/corr/abs-1711-07410, DBLP:journals/corr/abs-1910-10223}, however, as stated in \cite{DBLP:journals/corr/abs-1711-11586}, the quality improvement may come with the cost of losing multi-modality. Recent works show that it is possible to prevent losing multi-modality and use this method for multi-domain scenarios \cite{DBLP:journals/corr/abs-1711-11586, DRIT, huang2018munit}.

\textbf{Unsupervised disentanglement} aims to model the variations in data. It has been the focus of several pioneer works such as \cite{DBLP:journals/corr/ChenDHSSA16, Higgins2017betaVAELB, DBLP:journals/corr/abs-1811-11155}. InfoGAN \cite{DBLP:journals/corr/ChenDHSSA16}, for example, achieves this by maximizing the mutual information between latent variables and input data, whereas \cite{DBLP:journals/corr/abs-1811-05621, DBLP:journals/corr/abs-1711-11586, DRIT, DBLP:journals/corr/abs-2007-00653} disentangle input information to structure and texture codes. Our work builds on the same principles to disentangle structure and texture in a completely unsupervised approach. However, we go one step further and aim for better disentanglement by introducing a new module to enforce better separation between the two. We show that our approach can achieve the desired disentanglement and generate realistic and diverse images while disentangling structure from style better than previous methods.

\textbf{Multi-modal image synthesis} overcomes the limitation of conditional GANs ignoring the latent code, also known as mode collapse. The idea behind the multi-modal image-to-image translation is to learn a conditional distribution while generating diverse images. Early works on conditional image-to-image translation were mostly focused on producing deterministic outputs \cite{DBLP:journals/corr/IsolaZZE16, DBLP:journals/corr/LiuBK17},  which limits their applicability. In Section \ref{sec:experiments}, we show that our method can synthesize comparable results with the current state-of-the-art \cite{DBLP:journals/corr/abs-1711-11586, DBLP:journals/corr/abs-2003-12697}. 

\textbf{Style transfer} also known as texture transfer, can be defined as the problem of synthesizing an image with style extracted from the source image while preserving the semantics of the content image. Recent style transfer methods \cite{DBLP:journals/corr/abs-1812-04948, karras2020analyzing} proposed the use of conditional normalization layers such as Conditional Instance Normalization \cite{DBLP:journals/corr/DumoulinSK16} and Adaptive Instance Normalization \cite{DBLP:journals/corr/HuangB17} as a practical approach to transfer the global style. Normalization layers used in most style transfer methods diminish semantic information. Spatially-Adaptive Normalization \cite{park2019SPADE} was introduced as a way to avoid semantic-level information loss. We propose a closely related method for preserving semantic information without having access to a segmentation mask.

\vspace{-15pt}
\section{Method}
\vspace{-10pt}
Deep image manipulation requires an architecture with excellent feature extraction capabilities that allows for better disentanglement of texture from structure later on. Using an encoder, our goal is to disentangle the structure from the texture for both input images to our model. When swapping the texture or structure codes between the two randomly sampled input images $x_1, x_2 \in \mathbb{R}^{H \times W \times 3 }$, our model can synthesize an image with the same structural information as to its content reference, but having the visual appearance or texture of the style reference image. Thus, we aim to generate realistic synthesized images where the structure for the first image is preserved while transferring the style from the second image.

Our solution comprises three key modules with two discriminators namely $D$ and $D_{style}$ as shown in Figure \ref{fig:overview}: an encoder $E$, a generator $G$, and a disentanglement module $T$ which enforces better disentanglement of the structure from the style. The encoder learns how to encode visual information into two latent codes. Similar to \cite{DBLP:journals/corr/abs-2007-00653}, we enforce a mapping from any combination of the two latent codes to a realistic image by training an autoencoder. The generator synthesizes realistic images using the two extracted latent codes. The disentanglement module is designed to enforce the separation of the structure from the texture. We present the details of the objective function in the subsequent sections.

\vspace{-15pt}
\subsection{Encoder}
\vspace{-5pt}
The encoder $E$ learns a mapping from the input image to two latent codes corresponding to the structure and the texture. We use a traditional autoencoder training process. We employ a reconstruction loss to measure the difference between the original image and the synthesized version with an additional non-saturating adversarial loss \cite{goodfellow2014generative} to enforce realistic image generation, and is defined as,
\begin{equation} \label{eq_recon}
L_{enc}(x_1,\hat{x_1}) = L_{rec}(E,G) + L_{adv}(E,G,D) = \Vert x_1 - G(E(x_1)) \Vert_1 -\log(D(G(E(x_1))))
\end{equation}

\vspace{-20pt}
\subsection{Generator}
\vspace{-5pt}
Assuming we have already learned how to disentangle the structure from the texture, we can pass two images $x_1, x_2$ to the encoder and get the latent codes $z_1, z_2$ where $z_1=(z_s^1, z_t^1)$ and $z_2=(z_s^2, z_t^2)$. We assume $z_s$ is the encoded structure and $z_t$ is the texture of an input image and $\hat{x_1}$ is the reconstructed image. The generator conditioned on the latent structure code learns to map the extracted structure and texture codes to an image. The texture code will be added through weight modulation/demodulation introduced in \cite{karras2020analyzing}. Swapping the two texture codes before passing them to the generator is a common method to transfer style from one image to another. To ensure that the generated image is realistic, an additional non-saturating adversarial loss \cite{goodfellow2014generative} is added, given by,
\begin{equation} \label{eq_swap}
L_{swap}(E,G,D)  =  -\log(D(G(z_s^1,z_t^2)))
\end{equation}

\vspace{-20pt}
\subsection{Structure and texture disentanglement}
\vspace{-5pt}
The latent codes must represent the structure and texture. However, this cannot be achieved in our current setting without additional constraints to encourage consistent structure and texture disentanglement. The approach used for learning consistent texture codes is to enforce all the patches sampled from the image generated in the previous step by swapping the textures to be visually similar to patches extracted from the texture reference image \cite{DBLP:journals/corr/abs-2007-00653}. We achieve this using the following loss:
\begin{equation} \label{eq_style}
L_{style}(E,G,D_{style}) = -\log(D_{style}(C(G(z_s^1,z_t^2)), C(x^2))))
\end{equation}
where $C$ is a random crop of size in the range $[\frac{1}{8}, \frac{1}{4}]$. This formulation results in learning a more consistent style transfer.
Experiments have shown that this term is not enough and that better disentanglement can be achieved by enforcing the structure code not to contain texture-related information. In order to enforce structure consistency, we introduce an extra module with a gradient reversal layer as its first layer followed by a generator. Gradient reversal layer act as an identity function during forward but during backward it multiplies the gradients with $-1$. This new generator has the same architecture as the original generator, but it reconstructs an image with an all-zero texture code that is theoretically impossible. Our analysis of previous works shows that structure code contains spatial information and includes style-related information. An inconsistent encoding will cause the network to generate odd samples that do not follow the algorithms and cannot be interpreted. We train this module using a reconstruction loss and a non-saturating adversarial loss \cite{goodfellow2014generative}. 
\begin{equation} \label{eq_aux}
L_{aux}(x_1,\hat{x_1}) = L_{rec}(E,T) + L_{adv}(E,T,D) = \Vert x_1 - T(E(x_1)) \Vert_1 -\log(D(T(E(x_1))))
\end{equation}
Adding the gradient reversal layer, as shown in \cite{GRL}, forces the encoder to suppress any style-related information in the structure code. It also proved to be useful in cross domain disentanglement \cite{gonzalez2018image}. The auxiliary loss from this branch would help the encoder to disentangle structure from texture better.


\vspace{-10pt}
\subsection{Objective function}
\vspace{-5pt}
We jointly train the encoder, generators and discriminators to optimize the final objective, which is the weighted sum of previously mentioned loss functions and is given by,
\begin{equation} \label{eq_final}
L_{total} = \lambda_{rec} L_{enc} + \lambda_{swap} L_{swap} + \lambda_{style} L_{style} + \lambda_{aux} L_{aux} 
\end{equation}
where $ \lambda_{rec} , \lambda_{swap}, \lambda_{style}, \lambda_{aux} $ are weights that control the importance of each term. The optimal values used for each term are discussed in Section \ref{sec:experiments}.

\begin{table}
\centering
\begin{tabular}{@{}lccc@{}}
    \toprule
    Method & LSUN Church & \#iterations\\
    \midrule
    StyleGAN2 \cite{karras2020analyzing} & 57.54 & 48 M\\
    Swapping \cite{DBLP:journals/corr/abs-2007-00653} & 52.34 & 14 days x 4 V100 GPUs \\
    Ours(validation) & 51.42 & 5 M\\
    \bottomrule
    \end{tabular}
    \caption{Quantitative comparison of FID and training time/number of iterations on the validation set with state-of-the-art methods. Our proposed method achieves comparable performance while it converges significantly faster.} 
    \label{tab:sota}
\end{table}

\begin{figure}
    \centering
    \begin{subfigure}{0.47\textwidth}
        \includegraphics[width=\textwidth]{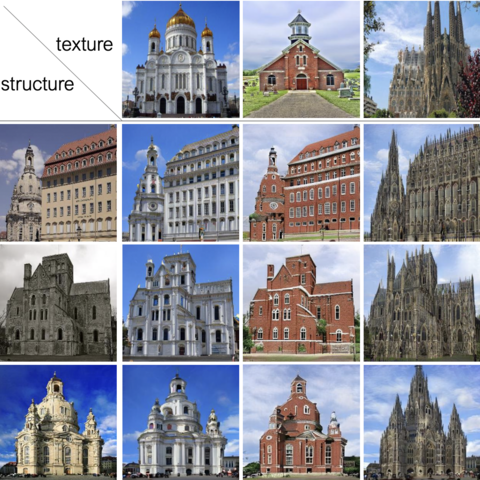}
        \caption{}
        \label{}
    \end{subfigure}
    \hfill
    \begin{subfigure}{0.47\textwidth}
        \includegraphics[width=.99\columnwidth]{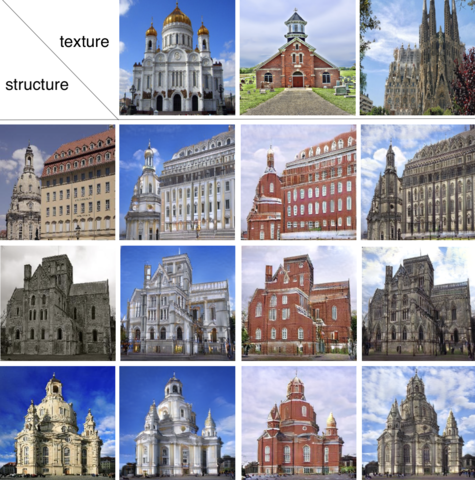}
        \caption{}
        \label{}
    \end{subfigure}
    \caption{Left: Results from Swapping Autoencoder \cite{DBLP:journals/corr/abs-2007-00653} on LSUN Church. Right: Our results on the same images. As evident, our model achieves better feature embedding and can retain the structural information of the input image while swapping only the texture with that of a second input image. Finer-level details such as spires and buildings outline are also retained. Most notably, our model was trained for a fraction of iterations compared to \cite{DBLP:journals/corr/abs-2007-00653}.}
    \label{fig:church_swapping}
\end{figure}

\vspace{-50pt}
\section{Experiments}
\label{sec:experiments}
\vspace{-10pt}
\noindent 
\textbf{Implementation details.} In all reported experiments, we randomly crop and resize the input images to $256\times 256$ resolution. We use the Adam optimizer \cite{kingma2014adam} with $\beta_1 = 0.0$, $\beta_2=0.99$. All reported results are computed on 4 NVIDIA TESLA P100 GPUs. The discriminator $D$ is based on StyleGAN 2 \cite{karras2020analyzing} and $D_{style}$ is based on Swapping autoencoder \cite{DBLP:journals/corr/abs-2007-00653}. 
We experimented with different hyper-parameters  for $ \lambda_{rec} , \lambda_{swap}, \lambda_{style}, \lambda_{aux} $ but in this version we simply set the loss weights to be all $1.0$.

\noindent
\textbf{Datasets.} We evaluate our method on four benchmark datasets curated for scene understanding and semantic segmentation.
\vspace{-5pt}
\begin{itemize}[leftmargin=*]
    \setlength\itemsep{0em}
    \item CelebAMask-HQ \cite{CelebAMask-HQ} has 30,000 face images collected from the CelebA \cite{liu2015faceattributes} dataset. CelebAMask-HQ contains annotations for 19 classes. However, we do not use masks in our training pipeline.
    \item LSUN church \cite{yu15lsun} is a subset of the Large-scale Scene Understanding (LSUN) dataset. The training set contains  126,227 images. It is a challenging dataset if no preprocessing is applied due to the diversity of the images. 
    
    \item Cityscapes \cite{DBLP:journals/corr/CordtsORREBFRS16} is a street view dataset collected from 50 cities across Germany. The training set contains 3000 images with fine annotations, and the test set contains 500 images. It is considered a challenging dataset for image-to-image translation because each scene may contain up to $30$ classes.   
    
    \item Inria \cite{maggiori2017dataset} is an aerial imagery dataset designed for semantic segmentation of building footprints. The training set contains 180 images with $5000 \times 5000$ resolution from 5 cities. Each image covers an area of approximately $1500 m \times 1500 m$. The test set contains 180 images of the same size collected from 5 cities that are not part of the training set. 
\end{itemize}

\noindent
\textbf{Baselines.} We compare our approach to a number of image-to-image translation, style transfer and multi-modal image synthesis methods including Swapping Autoencoder \cite{DBLP:journals/corr/abs-2007-00653}, StyleGAN2 \cite{karras2020analyzing} and BicycleGAN \cite{DBLP:journals/corr/abs-1711-11586}. We either use the results published by authors or generated using their official source code for all comparisons. 

\noindent
\textbf{Performance metrics.} We use Fréchet Inception Distance (FID) \cite{DBLP:journals/corr/HeuselRUNKH17} to measure the quality of generated images and LPIPS \cite{DBLP:journals/corr/abs-1801-03924} to compare the similarity of reconstructed images. FID calculates the difference between the real and the generated data distributions using the Inception network to extract the features while LPIPS calculates the perceptual similarity of the input with the reconstructed version. Additionally, in the Appendix(\ref{Appendix}), we report on the SIFID  metric on the LSUN church dataset for the training and testing sets, and include additional comparisons and use-cases.

\noindent
\textbf{Structure-consistent style transfer.} This section evaluates the quality of our generated images on style transfer and compares them to state-of-the-art. 
In Figure \ref{fig:church_swapping}, we provide a qualitative comparison of our synthesized images with our baselines. We find that our method produces comparable results with \cite{DBLP:journals/corr/abs-2007-00653} and \cite{karras2020analyzing} on LSUN Church dataset. A significant advantage of our approach is that it required only 5M iterations for training which demonstrates that not only is our approach significantly faster than our predecessors, but it surpasses their performance in terms of FID on the validation set, as shown in \ref{tab:sota}. Figure \ref{fig:church_swapping} shows that our method can generate samples with high visual quality on style transfer while preserving structure. 
Furthermore, structure similarity across generated samples supports the idea behind our auxiliary branch.

\begin{figure}[!ht]
    \centering
    \begin{subfigure}{0.47\textwidth}
        \includegraphics[width=\textwidth]{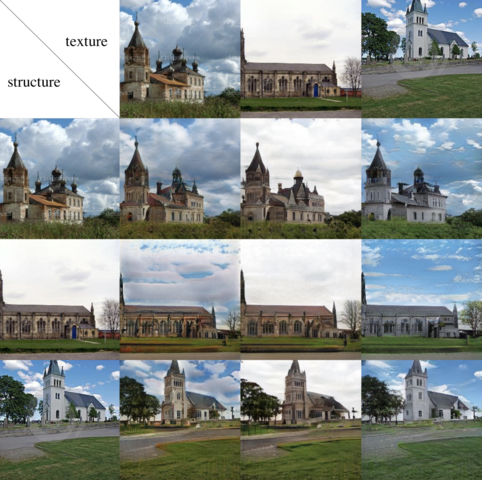}
        \caption{}
        \label{fig:church_leaf}
    \end{subfigure}
    \hfill
    \begin{subfigure}{0.47\textwidth}
        \includegraphics[width=\textwidth]{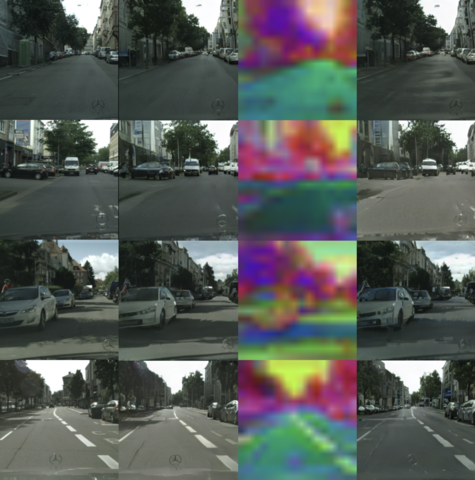}
        \caption{}
        \label{fig:cityscapes}
    \end{subfigure}
    \caption{(a) \textbf{Image translation on LSUN Church.} Each column corresponds to a particular texture extracted form the images on first row, respectively, each row contain the generated images with shared structure embedding. (b) \textbf{Image translation on Cityscapes.} The left column shows the input images from Cityscapes, the second column are reconstruction of input images. We provide a visualization of structure latent codes in the third column after applying PCA and then resizing it to $256 \times 256$ for the purpose of visualization. The last column shows our generated images by swapping the texture between first and third row and between second and fourth row. As it can be seen the lightning information, asphalt texture and coloring of the facades are the main information that transferred by swapping the texture codes.}
    
\end{figure}

\begin{figure}[!ht]
    \centering
        \begin{subfigure}[t]{0.47\textwidth}
        \includegraphics[width=\linewidth]{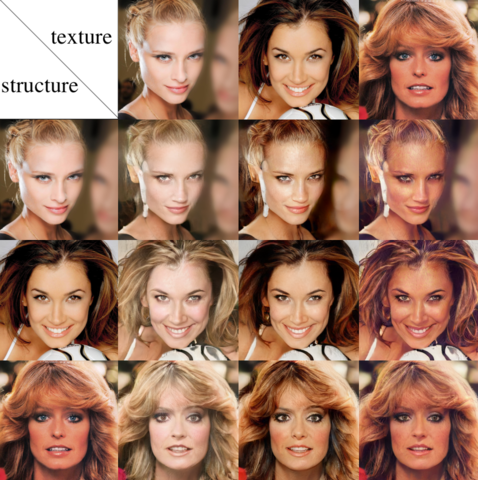}
        \caption{}
        \label{fig:celeba_2}
    \end{subfigure}
    \hfill
    \begin{subfigure}[t]{0.47\textwidth}
        \includegraphics[width=\textwidth]{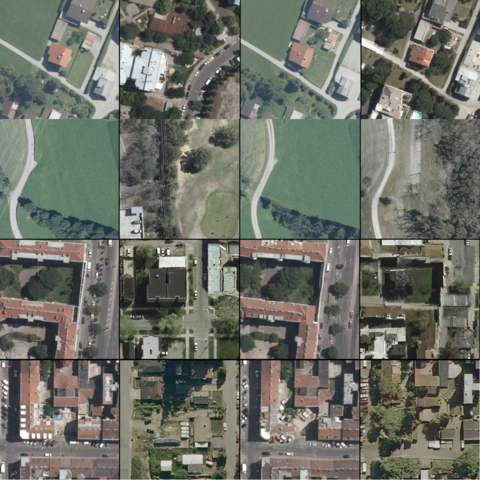}
        \caption{}
        \label{fig:inria}
    \end{subfigure}
    \caption{(a) \textbf{Style transfer on CelebAMask-HQ.} The first row shows the texture input image. The other rows show the results using the structure image in the first column. On the second row, the specular highlight on the face is embedded as a structure and is retained. 
    (b) \textbf{Performance on Inria dataset.} Left-to-right: first input $x_1$, second input $x_2$, reconstruction of $x_1$, our generated sample using structure of $x_1$ and texture of $x_2$. The semantic mask of $x_1$, if available, can be transferred to the synthetic image therefore increasing the labeled images in the training set that exhibit the textural characteristics of $x_2$.
    }
    \label{fig5}
\end{figure}

\noindent
\textbf{Realism of reconstruction.} The diagonals of Figure \ref{fig:celeba}, \ref{fig:celeba_2} and \ref{fig:church_leaf} show the quality of our method on image reconstruction task from the learned feature embedding. Our method preserves windows, doorways, trees, spires and generally the geometry of the objects as well as finer details such as earrings and tank top strap in Figure \ref{fig:celeba} (second row). We report quantitative comparison using the LPIPS \cite{DBLP:journals/corr/abs-1801-03924} to compare the similarity of reconstructed images. 

\noindent
\textbf{Disentanglement of structure and texture.}
Accurately disentangling structure and texture is an important task both for style transfer and image manipulation. Given that this disentanglement is performed entirely unsupervised, we can evaluate the effectiveness of our new module by comparing the performance of our method with previous works on style transfer from existing images. Better disentanglement of structure and texture leads to a finer manipulation, resulting in significantly more realistic images. Figure \ref{fig:church_swapping} (left) shows the results from Swapping Autoencoder \cite{DBLP:journals/corr/abs-2007-00653} on LSUN Church. Our results, shown on the right, demonstrate that our model achieves better feature embedding and generates images that retain the structural information of the input image while transferring only the texture from the second input image. Finer-level details such as spires and buildings' outlines are also preserved. 

\noindent
\textbf{Texture code normalization.}
We evaluated the effect of normalization on the texture latent code and found that applying $\mathcal{L}_{2}$-norm results in faster convergence and more realistic synthesis. In this work we do not employ normalization in the generator, as in \cite{DBLP:journals/corr/IoffeS15, DBLP:journals/corr/UlyanovVL16}, and similar to \cite{DBLP:journals/corr/abs-2007-00653}.
\noindent
\textbf{Contexts.}
In Figure \ref{fig:church_leaf}, we show examples from LSUN Church \cite{yu15lsun} that showcase the applicability of our method to other contexts. The bottom row shows a concrete example of how our technique preserves structures while transferring fine details. As it is evident, the building's structure is preserved while the texture is replaced. Similarly, the tree's structure is preserved, and its texture -in this case, the foliage's colour and density- changes according to each of the source images appearing in the top row. It should be noted that the model was not trained on any season transfer dataset. 
Semantic image synthesis is one of the critical tasks in designing 3D environments, image colorization, and image editing, but it requires semantic masks and corresponding input images for training a model. This poses a limitation for many real-world applications where it is not simple to produce segmentation masks to train a conditional generative model in a supervised setting, but they need accurate semantic consistency. Our method can perfectly adopt for semantically multi-modal image synthesis in an unsupervised setting.




\vspace{-15pt}
\subsection{Comparison to state-of-the-art}
\begin{figure}[!ht]
    \centering
    \begin{subfigure}{0.47\textwidth}
        \includegraphics[width=\textwidth]{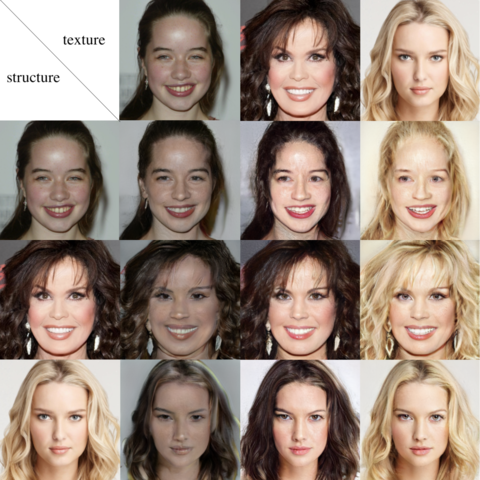}
        \caption{}
        \label{fig:celeba_3}
    \end{subfigure}
    \hfill
    \begin{subfigure}{0.47\textwidth}
        \centering
        \includegraphics[width=\textwidth]{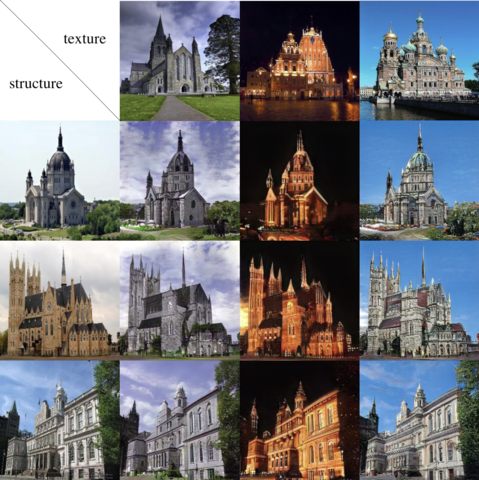}
        \caption{}
        \label{fig:church_sup}
    \end{subfigure}
    \caption{(a) Examples of style transfer on CelebAMask-HQ using our learned embedding. (b) Image translation on LSUN Church showing the quality of our method in different lightning and weather.}
\end{figure}

\vspace{-5pt}
Figure \ref{fig:celeba_3}, \ref{fig:church_sup}, and \ref{fig:inria_road} shows additional qualitative results on both reconstruction and style transfer tasks. The tables in Figure \ref{tab:sota_v2_validataion} and \ref{tab:ablations} present a quantitative comparison of our method with that of Swapping Autoencoder \cite{DBLP:journals/corr/abs-2007-00653}, StyleGAN2 \cite{karras2020analyzing}, MaskGAN \cite{CelebAMask-HQ}, and BicycleGAN \cite{DBLP:journals/corr/abs-1711-11586}. 

\begin{figure}
    \centering
    \begin{subfigure}{0.57\textwidth}
        \resizebox{\textwidth}{!}{ 
        \begin{tabular}{@{}lccc@{}}
        \toprule
        Method & LSUN Church & CelebAMask-HQ & Cityscapes  \\
        \midrule
        Ours & 51.42 & 29.69 & 162.46 \\
        Swapping \cite{DBLP:journals/corr/abs-2007-00653} & 52.34 & 32.83 & 182.5 \\
        StyleGAN2 \cite{karras2020analyzing} & 57.54 & - & - \\
        MaskGAN \cite{CelebAMask-HQ} & - & 46.84 & -\\
        BicycleGAN \cite{DBLP:journals/corr/abs-1711-11586} & - & - & 87.74\\
        \bottomrule
        \end{tabular}
        } 
        \caption{
        } 
        \label{tab:sota_v2_validataion}
    \end{subfigure}
    \hfill
    \begin{subfigure}{0.4\textwidth}
        \resizebox{\textwidth}{!}{ 
        \begin{tabular}{@{}lccc@{}}
        \toprule
        Method & LSUN Church  \\
        \midrule
        StyleGAN2 \cite{karras2020analyzing} & 0.377 \\
        Image2StyleGAN \cite{Abdal_2019_ICCV} & 0.186 \\
        Swapping \cite{DBLP:journals/corr/abs-2007-00653}& 0.227 \\
        Ours & 0.203 \\
        
        \bottomrule
        \end{tabular}
        } 
        \caption{} 
        \label{tab:ablations}
    \end{subfigure}
    \caption{(a) Quantitative comparison of FID on style transfer with some label-to-image translation work that are known for multimodal image synthesis and Swapping Autoencoder. In cases that we didn't have access to metric values calculated by the author, we trained their model for the same number of iterations as our network. Our method can achieve better results on CelebAMask-HQ and comparable results on LSUN Church trained for only 1.2M and 5M images. (b) Comparison of reconstructed image quality using LPIPS\cite{DBLP:journals/corr/abs-1801-03924} on LSUN Church. Our method focus on preserving structural details and can produce high quality results. Given the fact that our model have only been trained on 5M images which reduce the training time by a great factor, our method can reconstruct input images better than StyleGAN2 \cite{DBLP:journals/corr/abs-2007-00653}.}
    \label{fig:my_label}
\end{figure}

\vspace{-15pt}
\section{Applications}
\vspace{-10pt}
As stated earlier, an important motivation of our work is to remove biases from training datasets caused by class imbalances. Benchmark datasets such as \cite{liu2015faceattributes, DBLP:journals/corr/CordtsORREBFRS16} have inherent biases that adversely affect the network's generalization and significantly limit the effectiveness of networks used in real-world scenarios.

In this section, we present results on two unique applications employing the proposed technique: 
\begin{itemize}
    \item The first application addresses bias in training datasets and demonstrates how our method contributes to overcoming this issue.
    \item The second application addresses the cost-effective generation of training datasets for the task of semantic segmentation in satellite images without incurring additional labelling costs.
\end{itemize}
Furthermore, we present additional comparisons with state-of-the-art and quantitative results on the datasets LSUN Church \cite{yu15lsun}, CelebAMask-HQ \cite{CelebAMask-HQ}, Inria \cite{maggiori2017dataset}. We conclude with a discussion on the limitations of our technique.

\vspace{-15pt}
\subsection{Addressing bias in training datasets}
\begin{figure}[!ht]
\begin{center}
\includegraphics[width=\textwidth]{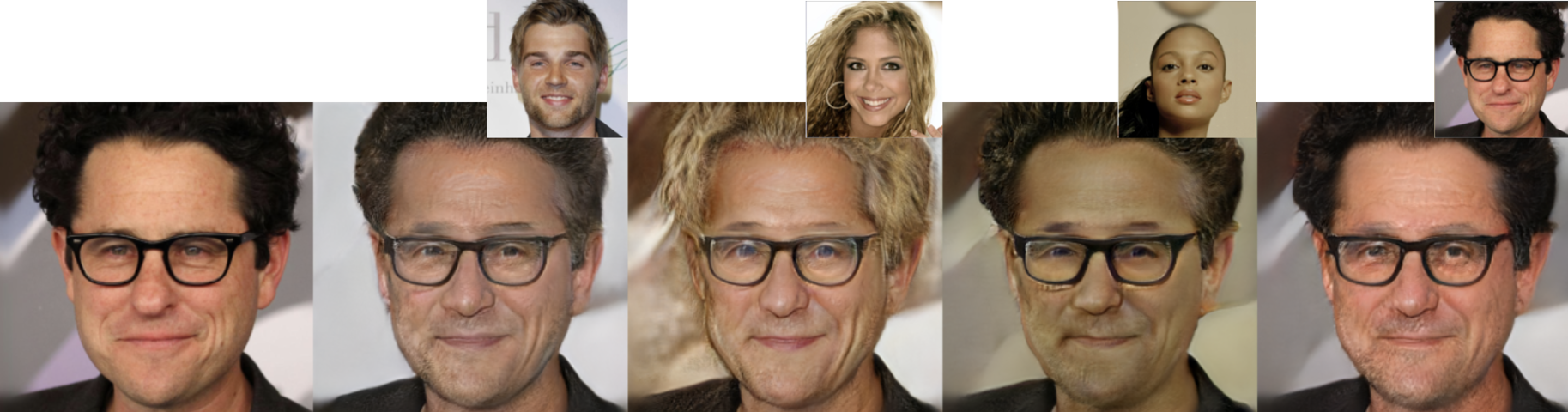}

\end{center}
\caption{
The first(left) image shows the first input image, and the second/third/fourth images show the generated image where the structure is retained from the first input image and the texture from the second/third/fourth input image, which appear in the inset images.}
\label{fig:celeba_sup}
\end{figure}

\vspace{-5pt}
Often we talk about biases in different datasets as an issue that needs to be addressed while designing the method, and we observe some generalization issues caused mainly due to imbalances in class distributions. A different approach is to adjust or expand our existing datasets to overcome this issue. Our method can preserve fine details; for example, in face datasets, these often imbalanced features can be gender, age, skin colour, hair colour, and accessories such as earrings, eyeglasses, hats, etc. Using our method allows us to balance the dataset by generating synthetic images with under-represented features. Furthermore, in cases where labels are available for the source image, these will also be the same for the generated images since our method preserves the same structure as the source image and only changes the appearance, as shown in Figure \ref{fig:celeba_sup}.

\begin{figure}[!ht]
]\centering
\includegraphics[width=\textwidth]{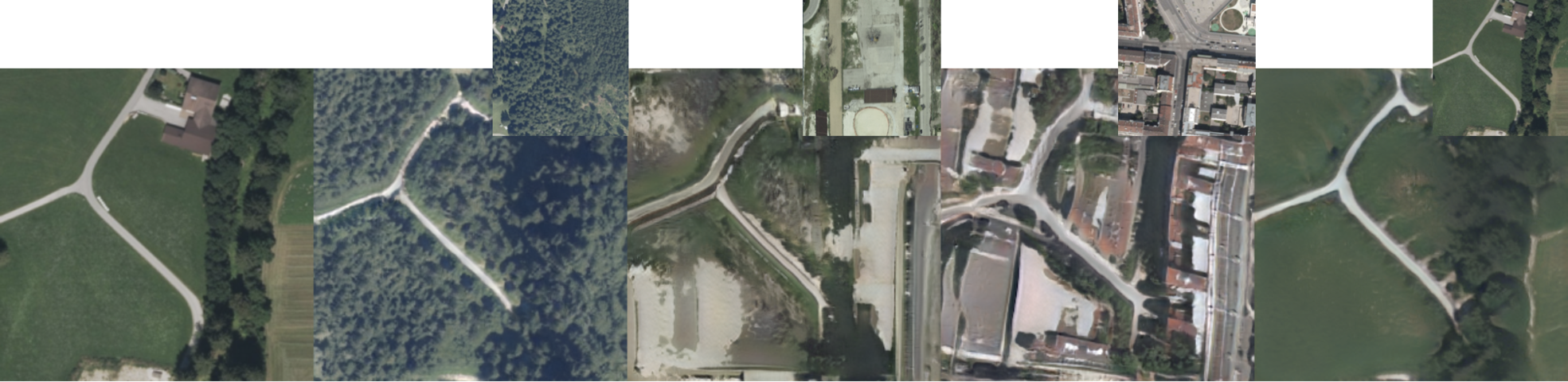}
\caption{
This figure provide an example of how our method can preserve the geometry of objects and semantic details while transferring the style. This would allow us to generate multiple samples with no extra labeling cost.
}
\label{fig:inria_road}
\end{figure}

\vspace{-15pt}
\subsection{Training datasets for Semantic Segmentation of Satellite Images} 
\vspace{-5pt}
Collecting satellite imagery for semantic segmentation is known to be an expensive and challenging task. The process of capturing images is expensive, but it may also contain inaccuracies due to the dynamic environment, e.g. a new building may appear that was not present at the time of acquisition of the satellite images. Another common issue is that the data collected from one city/continent cannot be easily generalized for a different city/continent. Considering all the challenges mentioned above, deploying a semantic segmentation network for aerial imagery can be challenging. Our structure-consistent network is designed to help overcome these challenges by generating realistic samples for different cities and weather conditions and generally creating datasets by style transfer. Our approach significantly reduces the time needed to process the data since we can expand any existing dataset to the desired style by only having a few images from the new city without requiring semantic labels Figure \ref{fig:inria_road}. Moreover, it can also be extremely useful for editing or expanding already existing datasets by changing the learned structure embedding.


\vspace{-20pt}
\section{Discussion and limitations}
\vspace{-10pt}
Our method is superior to state-of-the-art unsupervised approaches and gives comparable results to supervised techniques for image manipulation and image-to-image translation. We showed that incorporating the proposed auxiliary module as part of the training encourages better disentanglement of the structure from the texture and better feature embedding. This opens up new applications for image editing and style transfer, such as balancing existing datasets by generating images from underrepresented classes, expanding semantic segmentation datasets, creating multi-view datasets, etc. Previous works \cite{Dosovitskiy2020You} explored the effect of combining multiple loss functions with different weights in a single model using \cite{Higgins2017betaVAELB} to achieve better optimization. We believe the same can be applied as a future step on our pipeline for image manipulation. The importance of structure versus texture may differ from one application to another. By designing an architecture in which one can specify the percentage of structure versus texture for image generation, our method can address even broader range of challenges.

The proposed method works best when both structure and texture reference images contain the same object classes. Otherwise, the model's behaviour is not entirely predictable. An example of this limitation is where the texture reference image does not have vegetation, but the structure reference image contains a tree. In this scenario, the network may choose to copy the original texture. Additionally, in some cases, our network will generate an image with very little change to the structure image or replace some objects due to inconsistency between represented classes in the structure and texture reference images. We have not removed such cases during training. Ignoring them can be a reasonable next step for style transfer tasks until we better understand the underlying meaning of learned texture embedding. 

\vspace{-30pt}
\section{Conclusions}
\vspace{-10pt}
We presented an end-to-end process for training a structure-consistent image manipulation of existing images. We showed that our approach could disentangle structure and texture with higher accuracy while preserving finer details than state-of-the-art. We have extensively tested our method and showed that it could consistently transfer texture to the correct parts and preserve structural information without requiring a semantic mask. Most notably, this is achieved while also reducing the computational time needed for training such a network to a fraction of the time needed for the current state-of-the-art. Although our method outperforms much state-of-the-art in the image-to-image translation task, defining and disentangling structure from texture in multi-object scenarios such as Cityscapes remains challenging due to the diversity of the objects and complexity of the scene. In the future, we plan to explore the knowledge embedded in latent codes for different datasets and extend this framework to other domains as discussed in Section \ref{sec:experiments}. 

\section*{Appendix} 
\label{Appendix}
In this section we include additional qualitative and quantitative results, along with additional experiments, and a number of use-cases.

\section*{A. Use-case 1: Structure-guided style transfer}
Image-to-image translation and image synthesis can be used for various applications requiring finer control, such as multi-view image synthesis, expanding semantic segmentation datasets to increase their variability, addressing bias in existing datasets, etc. Generating images with more control over the structure and texture can potentially address problems where collecting data is costly or where bias is present in the training set.  

\begin{figure}[!ht]
	\centering
	\includegraphics[width=\linewidth]{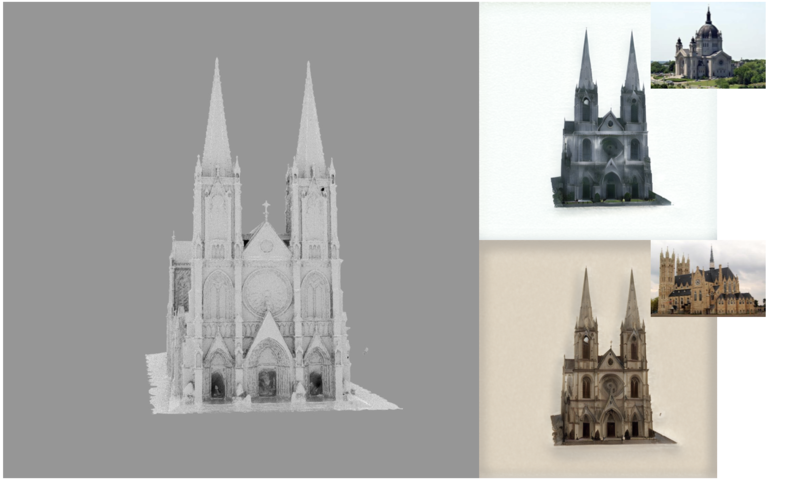}
	\caption{\textbf{What is structure?} In this experiment we used the image on the left of size 1024x1024 from a 3D mesh with no texture to better show case what is being transferred as structure.}
	\label{fig:1024}
\end{figure}

To verify the validity of the method and study the importance of the structure in image synthesis, we present the results in an extreme scenario where the source image used for structure does not contain texture information. To achieve this, we have used flat-shaded renders of the geometry as shown in Figure \ref{fig:1024}. The hypothesis is that given an image of the structure of the object, without any texture information, the method should preserve the fine details of the structure and transfer the related textures to the part of the image containing the structure of the object \textbf{only}. Figure \ref{fig:1024} shows the flat-shaded render of the scene geometry without any texture as structure input and the generated images after transferring the texture information from the inset images. As shown, the proposed method successfully preserves the detailed geometry and structure while transferring the style. It should also be noted that the only area in the generated image where texture is transferred is the part of the image corresponding to the scene geometry/structure, thus experimentally proving that the structure and texture codes are disentangled. In other words, if the structure tensor was not properly separated from the texture tensor for both source images, the texture would be transferred to other areas in the image.

\subsection*{B. Use-case 2: Differentiable Rendering}
\begin{figure}[!ht]
	\centering
	\includegraphics[width=\textwidth]{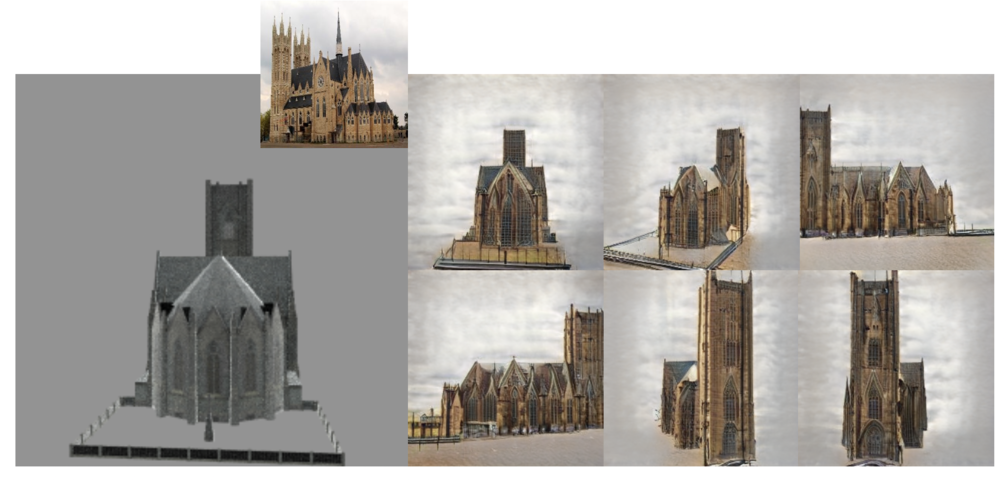}
	\caption{Generating multi-view imagery by transferring the texture from the inset image.}
	\label{fig:multiview_1}
\end{figure}

\begin{figure}[!ht]
	\centering
	\includegraphics[width=0.95\textwidth]{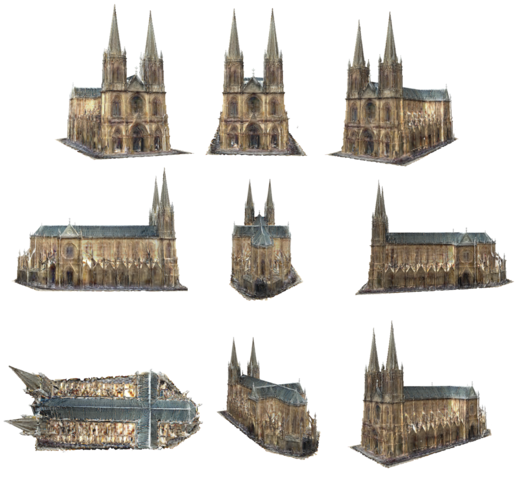}
	\caption{Renders using the spatially varying BRDF using a differentiable renderer on our multi-view images.}
	\label{fig:dif_renderer_2}
\end{figure}

\begin{figure}[!ht]
	\centering
	\includegraphics[width=\textwidth]{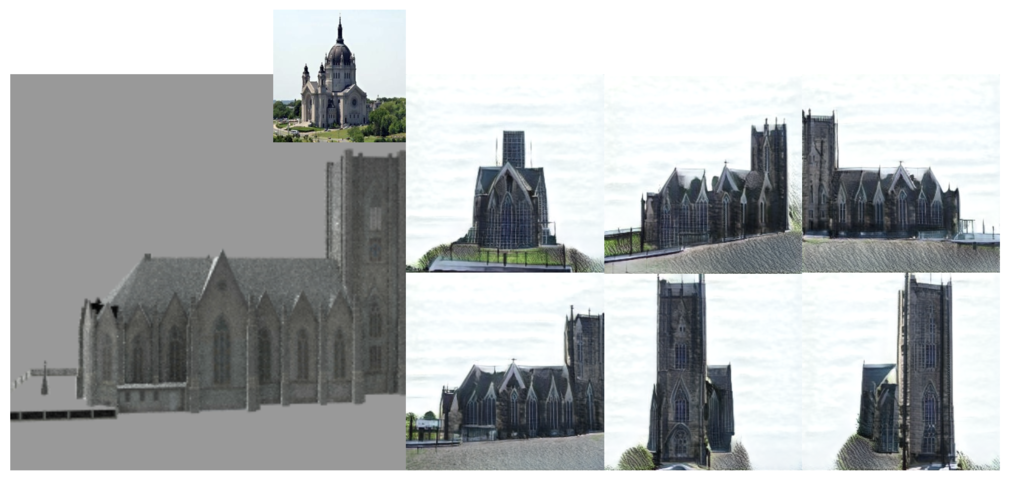}
	\caption{Generating multi-view imagery by transferring the texture from the inset image.}
	\label{fig:multiview_2}
\end{figure}

\begin{figure}[!ht]
	\centering
	\includegraphics[width=0.95\textwidth]{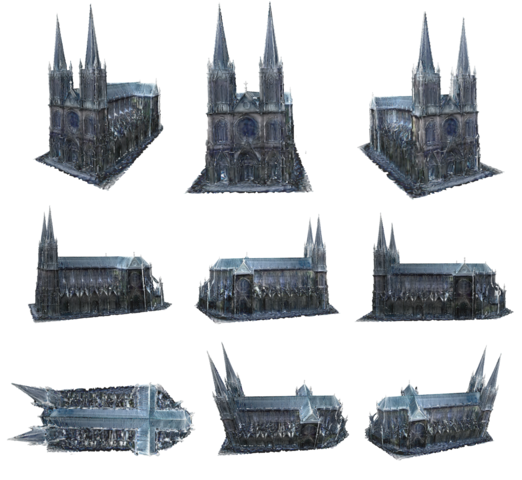}
	\caption{Renders using the spatially varying BRDF using a differentiable renderer on our multi-view images.}
	\label{fig:dif_renderer_1}
\end{figure}

To showcase the ability of our method ability to retain structure consistency, we perform experiments on generating multi-viewpoint imagery for differentiable rendering. Using our method, we first generate multi-view images by transferring any style to the desired structure as shown in Figures \ref{fig:multiview_1}, \ref{fig:multiview_2}. Next, we use a differentiable renderer to optimize the spatially varying BRDF(SVBRDF) properties from our generated multi-view images. Finally, we render using a Monte Carlo pathtracer to create renders of the geometry with the recovered SVBRDFs. Figure \ref{fig:dif_renderer_2} and \ref{fig:dif_renderer_1} show renders of the final results. As shown, the generated multi-view images using the proposed method preserve consistency and exhibit temporal coherence, which allows the application of differentiable rendering.


\clearpage

\section*{C. Additional Results}
Below, we present further quantitative and qualitative results.
\subsection*{C.1. Quantitative Results} 
\begin{figure}[h]
	\centering
	\includegraphics[width=0.4\linewidth]{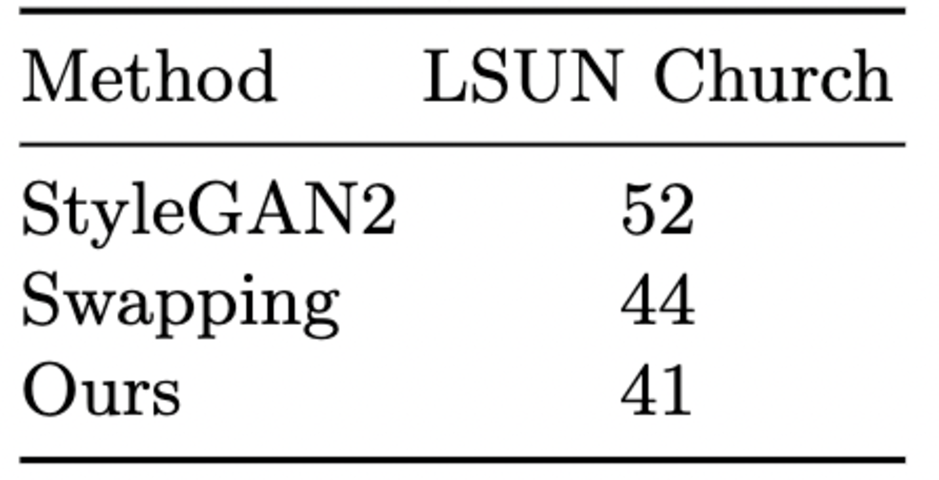}
	\caption{Comparison of the performance based on \textbf{Single-Image FID}\cite{rottshaham2019singan}}
	\label{fig:SIFID}
\end{figure}

In Figure \ref{fig:SIFID} we compare the texture similarity of images generated by our proposed  method with real samples based on SIFID\cite{rottshaham2019singan}.

\subsection*{C.2. Qualitative Results}
\begin{figure}
	\centering
	\includegraphics[width=\textwidth]{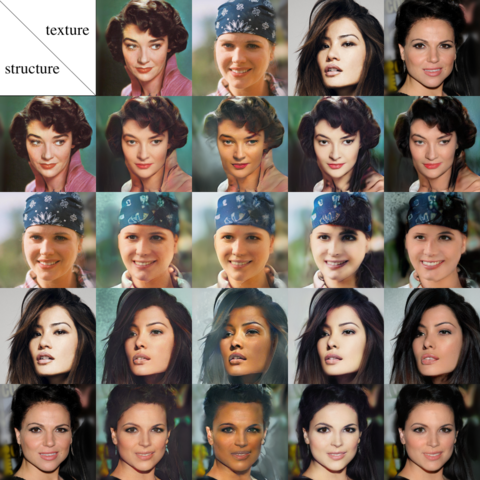}
	\caption{Randomly selected samples from CelebAMask-HQ.}
	\label{fig:celeba_5}
\end{figure}

\begin{figure}
	\centering
	\includegraphics[width=\textwidth]{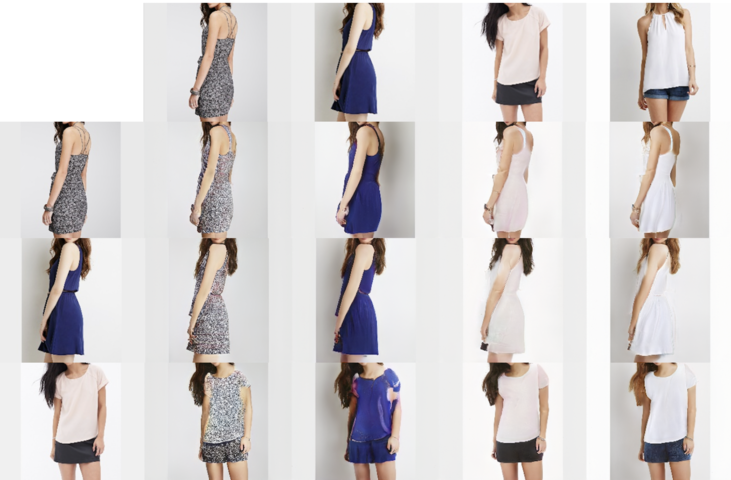}
	\caption{Randomly selected samples from the DeepFashion dataset. Some failure cases are shown in cases where the structure is not well-defined.}
	\label{fig:fashion}
\end{figure}

We show additional images generated by our method in Figure \ref{fig:celeba_5}. Our method can be used to expand training datasets used for semantic segmentation by increasing the variability and reducing bias by balancing the per-class training samples, e.g. complexion, hair characteristics (color, length, style), accessories (e.g. glasses, earrings, headbands, etc.), eye color, etc. as shown in Figure \ref{fig:celeba_5}. Similarly, Figure \ref{fig:fashion} shows the application of our method on the DeepFashion dataset. In these examples, we show some of the failures in cases where the structure is not well defined.

%
%
\clearpage
\bibliographystyle{splncs04}
\bibliography{samplepaper}
\end{document}